# Long-context Non-factoid Question Answering in Indic Languages

**Ritwik Mishra**[1] , **Rajiv Ratn Shah**[1] and **Ponnurangam Kumaraguru**[2]
[1] Indraprastha Institute of Information Technology, Delhi
{ritwikm, rajivratn}@iiitd.ac.in
[2] International Institute of Information Technology, Hyderabad
pk.guru@iiit.ac.in

## Abstract

Question Answering (QA) tasks, which involve extracting answers from a given context, are relatively straightforward for modern Large Language Models (LLMs) when the context is short. However, long contexts pose challenges due to the quadratic complexity of the self-attention mechanism. This challenge is compounded in Indic languages, which are often low-resource. This study explores context-shortening techniques, including Open Information Extraction (OIE), coreference resolution, Answer Paragraph Selection (APS), and their combinations, to improve QA performance. Compared to the baseline of unshortened (long) contexts, our experiments on four Indic languages (Hindi, Tamil, Telugu, and Urdu) demonstrate that context-shortening techniques yield an average improvement of 4% in semantic scores and 47% in token-level scores when evaluated on three popular LLMs without fine-tuning. Furthermore, with fine-tuning, we achieve an average increase of 2% in both semantic and token-level scores. Additionally, context-shortening reduces computational overhead. Explainability techniques like LIME and SHAP reveal that when the APS model confidently identifies the paragraph containing the answer, nearly all tokens within the selected text receive high relevance scores. However, the study also highlights the limitations of LLM-based QA systems in addressing non-factoid questions, particularly those requiring reasoning or debate. Moreover, verbalizing OIE-generated triples does not enhance system performance. These findings emphasize the potential of context-shortening techniques to improve the efficiency and effectiveness of LLM-based QA systems, especially for low-resource languages. The source code and resources are available at https://github.com/ritwikmishra/IndicGenQA.

## 1 Introduction

Question Answering (QA) represents one of the foundational tasks in Natural Language Processing (NLP) (Green et al., 1961). The objective of QA is to automatically generate or extract an appropriate answer to a given question.

Studying QA is crucial because learning through question-answer interactions is an effective method to train models in understanding nuances across various modalities, including text (Misra et al., 2018), video (Li et al., 2022), and audio (Fayek and Johnson, 2020). This work specifically focuses on the text modality.

Factoid questions, which can be answered with short spans of text, are a well-studied area in QA (Jurafsky, 2000). For example, the question "*Where is the Taj Mahal located?*" is an example of a factoid question. Recent advancements in QA systems have shown that factoid questions can often be addressed with near-human performance (Cortes et al., 2022). Moreover, Weissenborn et al. (2017) demonstrated that answering factoid questions can frequently rely on simple lexical heuristics between the question and its associated context.

This work emphasizes on non-factoid questions which require detailed and descriptive responses. For instance, a non-factoid question might be, "*How was the construction of the Taj Mahal perceived by the citizens of Agra in the 17th century?*" Despite the importance of such questions, non-factoid QA has received relatively less attention since the early stages of QA research (Surdeanu et al., 2008). Non-factoid QA systems are essential for applications such as generating responses for voice assistants like Amazon Alexa (Hashemi et al., 2020). Additionally, non-factoid questions are prevalent on web forums (Jijkoun and de Rijke, 2005), and search queries requiring long descriptive answers are commonly observed in Bing searches (Bajaj et al., 2016).

Non-factoid questions open avenues for the development of language models capable of processing long sequences (Soleimani et al., 2021). These questions are characterized by answers that span multiple passages; however, limited research has addressed this phenomenon comprehensively (Cortes et al., 2022). Transformer-based models are widely used in building QA systems (Garg et al., 2020; Nassiri and Akhloufi, 2023). Nevertheless, a notable limitation of Transformers is their quadratic memory complexity in the self-attention mechanism, which constrains their application in domains requiring the handling of longer sequences (Tay et al., 2021). While multilingual models such as mT5 and mBART exist, they inherit the same limitations as their English counterparts, particularly in managing long sequences effectively (Uthus et al., 2023). Although modern Large Language Models (LLMs) can process contexts with up to a million tokens (Team et al., 2024a), their performance in long-context QA tasks often deteriorates, especially when the answer is not located near the beginning or end of the context (Liu et al., 2024a). In this study, we investigate the use of Answer Paragraph Selection (APS) models, Open Information Extraction (OIE), and coreference resolution to reduce the context length associated with non-factoid questions.

Open Information Extraction refers to the task of extracting n-ary tuples from a given text. When these n-ary tuples consist of three components—namely, the head, relation, and tail—they are referred to as triples. For instance, consider the sentence: *Helen, John Wick's wife, gifted him a beagle puppy named Daisy*. The extracted triples from this sentence might include: (i) {*Helen*, wife-of, *John Wick*}, (ii) {*Helen*, *gifted*, *him*}, and (iii) {*Daisy*, is-a, *beagle*}. Notably, triple extraction is not a deterministic process, and different sets of triples may be derived from the same text. Coreference resolution, on the other hand, involves identifying text spans that refer to the same real-world entity. For example, in the aforementioned sentence, the terms *him* and *John Wick* both correspond to the same entity.

Similar to many subfields in computational linguistics, the majority of research on Question Answering (QA) has been conducted in English (Soares and Parreiras, 2020). However, (Sagen, 2021) demonstrated that utilizing model architectures designed for handling long contexts and pretrained on high-resource languages does not yield optimal performance for long-context scenarios in low-resource languages. For non-factoid QA systems, (Hu et al., 2020) revealed that models trained on English often achieve near-human performance. Furthermore, non-factoid QA datasets predominantly feature English, followed by Chinese, Arabic, European, and South Asian languages, while Indic languages remain underrepresented (Cortes et al., 2022). For our experiments, we focus on Indic languages such as Hindi, Urdu, Tamil, and Telugu, as publicly available open information extraction (OIE) tools exist for these languages (Mishra et al., 2023; Kolluru et al., 2022a).

The critical importance of non-factoid question answering (QA) systems and the limited research on Indic non-factoid QA have inspired us to develop models tailored for non-factoid QA in Indic languages. The primary contributions of this study are as follows:

1. We demonstrate that the performance of LLMs in non-factoid QA can be significantly improved by incorporating an Answer Paragraph Selection (APS) model to reduce the context size effectively.

2. We release the finetuned checkpoints for gemma-2b, gemma-7b, and Llama3.1, trained on non-factoid question-answer pairs in Hindi, Urdu, Tamil, and Telugu.

3. We propose a novel Semantic Text Similarity score for Multilingual Texts (STS-MuTe), specifically envisioned to evaluate the quality of generated answers.

4. Our analysis reveals that when the APS model assigns a high score to a question-paragraph pair, nearly all tokens contribute to this prediction. Conversely, when a low score is predicted, only a few tokens are influential in the APS model's decision.

The structure of this paper is organized as follows: Section 2 provides an overview of foundational studies in question-answering (QA) methodologies, with a specific focus on works addressing QA systems based on retrievers, open information extraction, coreference resolution, and non-factoid questions. Section 3 presents a concise description of the dataset employed in this study. Section 4 details the overall methodology adopted to design a QA pipeline, elaborating on the approaches used, implementation specifics, and evaluation metrics.

Section 5 reports the results obtained in this study. Section 6 examines the findings derived from post-hoc explainable AI methods applied to the APS model, highlighting the rationale employed by the model. This section also discusses the types of questions that were effectively addressed by the proposed QA pipeline. The paper concludes with Section 7, summarizing the work and presenting its limitations and future research directions, as elaborated in Sections 8 and 9, respectively.

## 2 Literature Survey

It has been observed that QA tasks involving context (either long or short) and typically yielding factoid-type answers are often referred to as Machine Reading Comprehension (MRC) (Liu et al., 2019). A cloze task represents a specialized form of QA, wherein no explicit question is provided, but the task requires filling in blanks based on provided instructions (Taylor, 1953).

In the absence of a universally accepted standard for defining a long context, this study adopts a threshold of 512 tokens to classify a context as long. The rationale behind selecting this threshold is twofold: (a) it corresponds to the default input limit of transformer-based encoder architectures, such as BERT (Devlin et al., 2019) and RoBERTa (Liu, 2019), which are extensively employed in question-answering (QA) models (Navarro et al., 2024); and (b) the context length in the dataset utilized in this study (refer to Section 3) exceeds this threshold. We discuss the token limit of LLMs in Section 8.

### 2.1 Retrievers

Li et al. (2024a) demonstrated that contemporary language models, such as Mamba, which are not based on the Transformer architecture, exhibit inferior performance compared to Transformer-based LLMs on tasks requiring long-context understanding. Similarly, Wang et al. (2022) investigated the task of open-domain non-factoid question answering, referred to as long-form question answering, and found that retrieval-based approaches outperform direct generative answering methods in this context. The utility of retrieval techniques for open-domain question answering was also highlighted by Karpukhin et al. (2020). Furthermore, Huo et al. (2023) posited that text retrieval can mitigate hallucinations in LLMs when addressing long-context questions in open-domain scenarios.

### 2.2 Open Information Extraction

Triples and knowledge bases have been extensively utilized in the literature to enhance question-answering pipelines (Litkowski, 1999; Berant et al., 2013). The authors of K-BERT (Liu et al., 2020) incorporated knowledge derived from OpenIE triples to address various tasks, including question answering. Similarly, Fader et al. (2014) employed both synthetic and gold-annotated triples from diverse knowledge bases to answer open-domain user queries. Studies such as Baek et al. (2023) and Sen et al. (2023) have demonstrated that retrieving knowledge base triples, verbalizing them, and incorporating them as context significantly improves the performance of LLMs. Furthermore, Khot et al. (2017) leveraged OpenIE triples alongside the tabular reasoning method proposed by Khashabi et al. (2016) to determine the correct answer for multiple-choice questions.

### 2.3 Coreference Resolution

Zhang et al. (2024) and Bai et al. (2024) highlighted the subpar performance of contemporary LLMs on English and Chinese question-answering (QA) tasks when the context length approximates that of a novel. However, Liu et al. (2024b) demonstrated that incorporating coreference resolution as a pre-processing step prior to query-specific short context retrieval significantly enhances the performance of LLMs, even for QA tasks involving novel-length contexts. Additionally, Chai et al. (2023) proposed a pipeline leveraging Open Information Extraction (OIE) triples and coreference resolution to address QA tasks. Their approach further employed post-hoc explainable artificial intelligence (XAI) techniques, including LIME (Ribeiro et al., 2016a) and SHAP (Lundberg and Lee, 2017a), to interpret the generated answers. However, this pipeline was not evaluated on any extensive dataset.

### 2.4 Non-factoid QA systems

The literature demonstrates that non-factoid questions can be addressed using FAQ databases (Jijkoun and de Rijke, 2005; Berger et al., 2000; Soricut and Brill, 2006). Agichtein et al. (2004) proposed a query transformation approach followed by document ranking to resolve non-factoid queries. However, their methodology does not provide a clear mechanism for extracting concise answers from the top-ranked documents. Similarly, Jijkoun and de Rijke (2005) and Surdeanu et al. (2008) adopted answer ranking as an initial step. Existing studies also reveal that semantic techniques, such as BM25, perform inadequately when applied to non-factoid questions (Cohen and Croft, 2016). Merely retrieving sentences or paragraphs from documents has been shown to be insufficient for effectively addressing non-factoid questions (Yang et al., 2016). Consequently, LLMs are essential for generating comprehensive answers.

Cohen et al. (2018) employed various APS techniques specifically to address non-factoid questions. Glass et al. (2022) demonstrated that the process of retrieving, re-ranking passages, and subsequently generating answers from them leads to improved performance. Yulianti et al. (2017) initially retrieved relevant answer paragraphs and then generated summaries from them. The approach of extracting a short context from a longer passage not only decomposes the long-context question answering problem into distinct steps but also enhances interpretability in the results (Caciularu et al., 2022). Li et al. (2024c) integrated a document ID retriever with document-grounded answer generation to propose a unified framework for generative question answering.

Singh et al. (2024) introduced a QA benchmark for Indic languages by consolidating all publicly available QuAD datasets in Indic languages and augmenting them with synthetic data. Nonetheless, the representation of non-factoid questions within multilingual QuAD datasets remains an area that requires further exploration.

The research gaps identified in this study are as follows: (a) A lack of long-context non-factoid question answering (QA) models capable of handling multilingual data. (b) Although OpenIE triples and coreference resolution models have demonstrated effectiveness in resource-rich languages such as English and Chinese, their applicability to low-resource languages remains insufficiently explored. (c) The application of post-hoc explainable AI (XAI) methods, such as LIME and SHAP, has yet to be thoroughly examined in retrieval-based QA systems. This work seeks to address these identified research gaps.

## 3 Dataset

In this study, we utilized the MuNfQuAD (Mishra et al., 2024b), the largest publicly available multilingual non-factoid question answering dataset. The question-answer pairs and their corresponding context were extracted from BBC news articles. The dataset is publicly available for research purposes with due permission from BBC. Approaches similar to those of (Soleimani et al., 2021) and (Hashemi et al., 2020), which employed web scraping techniques to obtain non-factoid question-answer pairs from the open web, have also been explored. We leveraged the finetuned APS model from the MuNfQuAD work, ensuring that the training data of the APS model was excluded from any finetuning or evaluation performed in this study. After excluding the APS model's training data from MuNfQuAD, the resulting dataset comprised over 40K question-answer pairs across four languages.

In our investigation of the potential of Open Information Extraction (OIE) and coreference resolution for long-context non-factoid question answering, we were constrained to languages for which such tools are available. The coreference resolution model of Transmucores (Mishra et al., 2024a) has been trained on 31 South Asian languages. However, there is a notable scarcity of multilingual OIE tools. To the best of our knowledge, IndIE (Mishra et al., 2023) and Gen2OIE (Kolluru et al., 2022b) are the only two multilingual OIE tools that can be used with four Indic languages: Hindi, Tamil, Telugu, and Urdu. Consequently, this study evaluates the proposed methodology on these four languages.

## 4 Methodology

A conventional method for addressing a query within a provided context involves presenting the question-context pair to a LLM. Empirical evidence suggests that LLMs exhibit superior performance compared to alternative models on various NLP tasks (Dubey et al., 2024; Team et al., 2024b; Labruna et al., 2024). The conventional approach to utilizing LLMs for question answering, contingent upon the given context, is visually depicted in Figure 1a.

In this work, we investigate the potential of using a retriever in the QA pipeline to reduce a long-context to a short-context to answer non-factoid questions. Figure 1b illustrates the retrieval method. As highlighted in section 2, prior methods have already proposed to use a retriever in QA pipeline. We aim to investigate its utility in multilingual QA pipeline with non-factoid questions. In this work, we used the following four approaches to build a retriever shown in Figure 1b.:

A1. **Simple APS model:** We pass the paragraphs

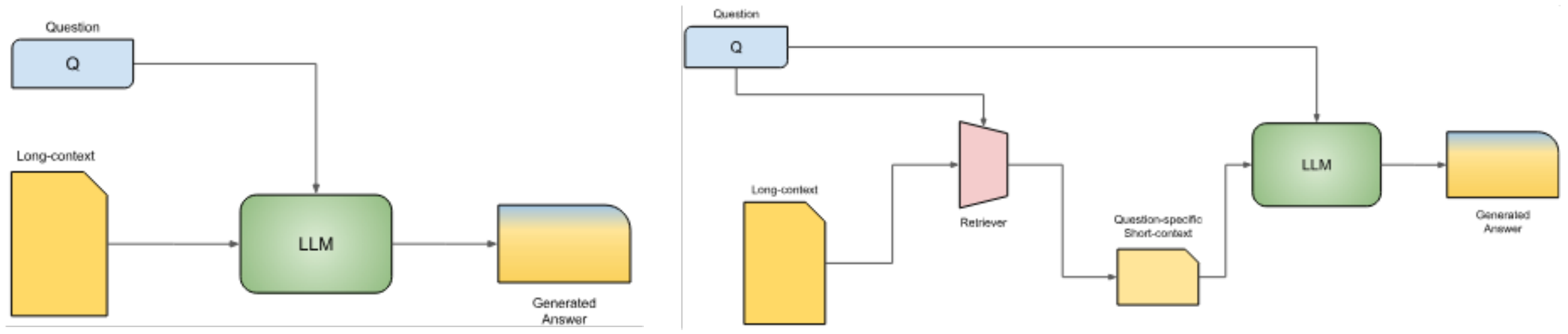

(a) Baseline method: Entire long-context is fed to the LLM with question text and answer is generated based on the given prompt instruction.

(b) Retrieval Method: A short-context is derived by the retriever using the long-context and the question text. This obtained short-context is subsequently provided as input to the LLM.

Figure 1: Methods for long-context question answering using Large Language Models (LLMs).

in long-context and question to the finetuned APS model of (Mishra et al., 2024b). Top-5 paragraphs were selected as the short-context.

A2. **Ranking verbalized triples (OIE+APS):** We extracted triples from each paragraphs of the given long-context. A triple is consisted of a head, relation, and a tail. During the verbalization phase, short sentences were generated by concatenating the extracted head, relation, and tail. Owing to the free word order characteristic of Indian languages (Tandon and Sharma, 2017), the resulting sentences retained their semantic coherence. We take the verbalized triples and concatenate it with the question to pass them to the APS model. Using the APS model output we rank and filter top-10 verbalized triples and use them as the short context.

A3. **Coreference resolution and APS ranking (coref+APS):** We run the multilingual coreference resolution from (Mishra et al., 2024a) on the long-context to get the coreference links between paragraphs. After selecting the paragraph which gives highest score by APS model for the given question, we filtered all the paragraphs that contains coreference links to this paragraph. Top-5 paragraphs are selected based on their APS scores.

A4. **Ranking coreference chains of verbalized triples (OIE+coref+APS):** Using the predicted coreference chains and verbalized triples, we constructed clusters of verbalized triples according to the coreference chains. Figure 2 shows the visual representation of the four approaches used in this work.

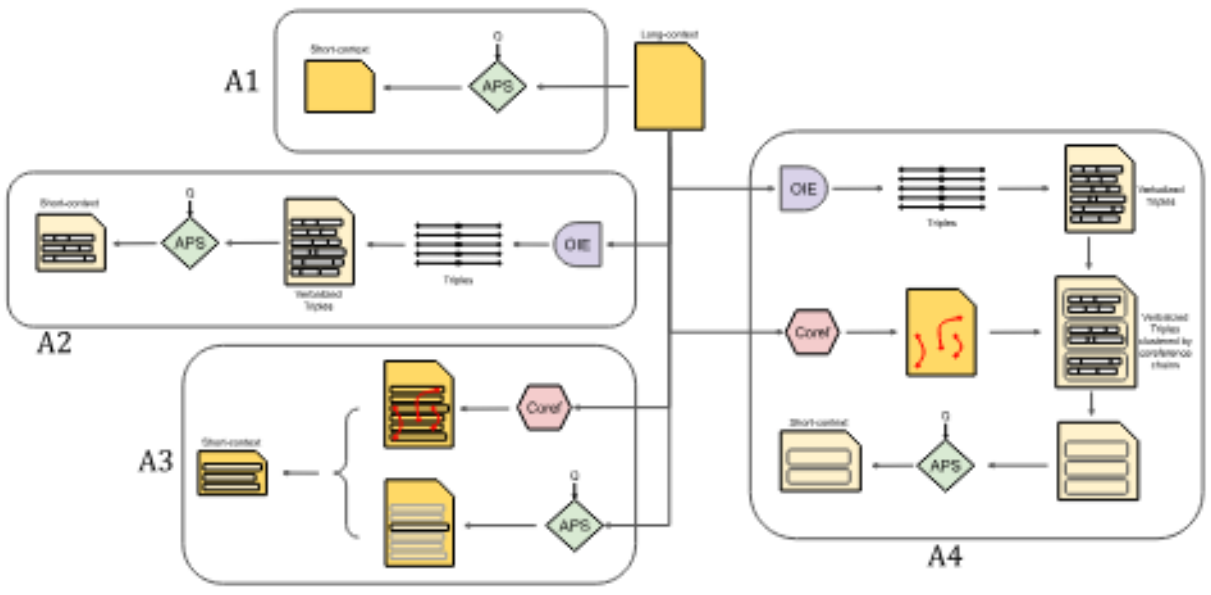

Figure 2: Illustration of the four approaches used to construct the retriever. The verbalization of triples refers to the process of transforming a triple (composed of a head, relation, and tail) into a short sentence by combining its constituent components.

### 4.1 Implementation

In this work, we use the finetuned APS model from (Mishra et al., 2024b) to predict a score that indicates whether the given text can answer the given question. Based on these scores, paragraphs are ranked, and the top five passages (k=5) are selected. It has been observed that simpler baselines like tf-idf and cosine similarity on word embeddings are not sufficient to perform effective answer paragraph retrieval from a given document with long-context (Keikha et al., 2014; Qu et al., 2019). Moreover, (Feng et al., 2015) has showed that deep learning architectures which are based on finetuning embeddings performs well for a answer paragraph retrieval.

We have chosen checkpoint with the base encoder of XLM-R (Conneau et al., 2020) since we have observed that XLM-R performed better on multilingual QA as compared to mbert (Hu et al., 2020). Morever, the GPU memory footprint and inference time on XLM-R based APS model is significantly less than other multilingual encoders.

We compare our approaches (A1-A4) with the baseline (B) shown in Figure 1a where no retriever

```
Answer the question based on the given context.
##Question
{question}
##Context
{context}
##Answer
{answer}
```

Figure 3: Prompt used to finetune LLMs in the QA pipeline. The {context} will be a long-context for QA pipeline with B whereas it will be a short-context for QA pipeline with A1-A4. During the inference stage, {answer} was omitted from the prompt.

is used in the QA pipeline. We used 2 billion and 7 billion parameter model of instruction finetuned Gemma (Team et al., 2024b). We also used 8 billion parameter model of instruction finetuned Llama 3.1 (Dubey et al., 2024). For each approach (A1-A4) and baseline (B), we finetuned the aforementiond LLMs on a training set of more than 29K questions. We used parameter efficient finetuning (Mangrulkar et al., 2022) with LoRA (Hu et al., 2021) rank and alpha of 32 to develope the *QA pipeline containing finetuned LLMs* ($QA_{finetuned}$). The prompt used for finetuning LLMs is shown in Figure 3. The selection of these models was motivated by two primary considerations: (a) their parameter sizes are compatible with the computational resources available to us, and (b) these models have demonstrated effectiveness in prior studies on question answering in low-resource languages (Trivedi et al., 2024; Al Nazi et al., 2025; Qiu et al., 2024).

We ran zero-shot inferences while running the models on the test set. Four bit quantization[1] was used to run inferences with *QA pipeline containing non-finetuned (base) LLMs* ($QA_{base}$). All the experiments were performed on two NVIDIA A100 cards with each card having 40GB of GPU RAM. In order to ensure reproducibility, we used low temperature value (0.001) for local LLMs and API calls.

### 4.2 Evaluations

In order to compare the LLM generated answers with silver answers of MuNfQuAD, we used ROUGE score (Lin, 2004) to quantify the similarity between the two answers. (Bajaj et al., 2016) also used ROUGE metric for evaluating long descriptive answers of a question. However, Soleimani et al. (2021) has shown that ROUGE metric cannot be trusted to evaluate long sequences. Furthermore, while LLM-generated answers may perform poorly on lexical metrics, they often convey the same semantic information as the ground truth answers (Kamalloo et al., 2023). Hence we propose a semantic text similarity score for multilingual texts (STS-MuTe). For a given text pair $(t_1, t_2)$, STS-MuTe is computed as the arithmetic mean[2] of the cosine similarity ($cos$) between various multilingual embeddings and the BERTScore (Zhang et al., 2020). We used the following models to calculate the cosine similarity: (a) USE (Cer et al., 2018), (b) LaBSE (Feng et al., 2022), and (c) LASER (Schwenk and Douze, 2017; Artetxe and Schwenk, 2019). BERTScore, which is based on token-level pairwise cosine similarity, produces a single floating-point value indicative of semantic similarity. For this purpose, we utilized pretrained multilingual BERT (Devlin et al., 2019) to generate contextual embeddings required for BERTScore computation. Therefore, in this work we use token-level and semantic-level similarity measures to evaluate the answers generated by LLMs.

## 5 Results

Our test set consisted of more than 6k questions. However, due to resource constraints, we had to limit the inference on a subset of 1100 questions from the test set. Average context length in these 1100 examples was 1070 tokens. We ensured that the four languages in consideration are all represented in this subset of the test set. Table 1 illustrates the performance achieved by our QA pipeline with different approaches. We ensured that the test samples on which $QA_{base}$ is evaluated are the same test samples on which $QA_{finetuned}$ is evaluated.

As compared to the baseline (B), it was observed that A1 outperformed other approaches on nearly all evaluation metrics. The instances where performance of B was shown to be slightly better than A1 (gemma-7b-it) even in those instances A1 was seen to be giving competitive values. It was observed that IndIE method performed better than Gen2OIE in A2 whereas Gen2OIE performed better in A4. We attribute this to the fact that IndIE has shown to generate more fine-grained triples than Gen2OIE (Mishra et al., 2023). Our results indicate that APS model is able to extract better short-context from verbalization of fine-grained triples. Whereas coarse triples of Gen2OIE are

[1]https://pypi.org/project/bitsandbytes/

[2]We also experimented with taking harmonic mean instead arithmetic mean. But no significant change in results was observed.

| | BERTScore | USE | LaBSE | LASER | STS-MuTe | R1 | R2 | R3 | RL |
|---|---|---|---|---|---|---|---|---|---|
| **Gemma 2 Billion Instructional (gemma-2b-it)** | | | | | | | | | |
| | Semantic-level | | | | | Token-level | | | |
| B | 0.63 (0.75) | 0.38 (0.80) | 0.43 (0.58) | 0.58 (0.83) | 0.50 (0.74) | 0.10 (0.44) | 0.04 (0.31) | 0.02 (0.29) | 0.08 (0.34) |
| A1 | **0.66 (0.78)** | **0.39 (0.81)** | **0.46 (0.70)** | **0.56 (0.85)** | **0.52 (0.79)** | **0.12 (0.51)** | **0.08 (0.44)** | **0.07 (0.42)** | **0.11 (0.39)** |
| A2 | 0.62 ( 0.61 ) | 0.31 ( 0.53 ) | 0.38 ( 0.51 ) | 0.50 ( 0.74 ) | 0.46 ( 0.60 ) | 0.07 ( 0.14 ) | 0.02 ( 0.04 ) | 0.01 ( 0.02 ) | 0.05 ( 0.11 ) |
| A3 | 0.57 (0.64) | 0.18 (0.61) | 0.32 (0.50) | 0.49 (0.77) | 0.39 (0.63) | 0.03 (0.21) | 0.01 (0.09) | 0.00 (0.07) | 0.03 (0.14) |
| A4 | 0.65 ( 0.75 ) | 0.37 ( 0.77 ) | 0.45 ( 0.67 ) | 0.55 ( 0.83 ) | 0.50 ( 0.75 ) | 0.11 ( 0.43 ) | 0.07 ( 0.35 ) | 0.06 ( 0.33 ) | 0.10 ( 0.33 ) |
| **Gemma 7 Billion Instructional (gemma-7b-it)** | | | | | | | | | |
| | Semantic-level | | | | | Token-level | | | |
| | BERTScore | USE | LaBSE | LASER | STS-MuTe | R1 | R2 | R3 | RL |
| B | 0.66 (**0.81**) | **0.52** **(0.84)** | 0.53 (0.72) | **0.70** **(0.87)** | 0.60 (**0.81**) | 0.16 (**0.55**) | 0.07 (**0.47**) | 0.05 (**0.46**) | 0.11 (**0.49**) |
| A1 | **0.69** (0.79) | **0.52** (0.81) | **0.56** **(0.74)** | 0.66 (0.86) | **0.61** (0.80) | **0.19** (0.53) | **0.13** (0.46) | **0.11** (0.44) | **0.16** (0.42) |
| A2 | 0.65 ( 0.63 ) | 0.44 ( 0.58 ) | 0.46 ( 0.56 ) | 0.59 ( 0.77 ) | 0.53 ( 0.63 ) | 0.10 ( 0.17 ) | 0.03 ( 0.06 ) | 0.01 ( 0.03 ) | 0.07 ( 0.12 ) |
| A3 | 0.63 (0.70) | 0.43 (0.69) | 0.45 (0.62) | 0.63 (0.81) | 0.54 (0.71) | 0.09 (0.33) | 0.03 (0.21) | 0.01 (0.17) | 0.07 (0.24) |
| A4 | 0.68 ( 0.76 ) | 0.49 ( 0.77 ) | 0.53 ( 0.70 ) | 0.63 ( 0.83 ) | 0.58 ( 0.77 ) | 0.16 ( 0.45 ) | 0.11 ( 0.36 ) | 0.09 ( 0.34 ) | 0.14 ( 0.35 ) |
| **Llama 3.1 8 Billion Instructional (llama3.1-8b-it)** | | | | | | | | | |
| | Semantic-level | | | | | Token-level | | | |
| | BERTScore | USE | LaBSE | LASER | STS-MuTe | R1 | R2 | R3 | RL |
| B | 0.69 (**0.78**) | 0.61 (0.77) | 0.57 (0.72) | 0.69 (**0.82**) | 0.64 (0.77) | 0.28 (0.46) | 0.18 (0.39) | 0.15 (0.38) | 0.22 (**0.41**) |
| A1 | **0.70 (0.78)** | **0.68 (0.78)** | **0.62 (0.73)** | **0.76 (0.82)** | **0.69 (0.78)** | **0.34 (0.49)** | **0.26 (0.43)** | **0.24 (0.41)** | **0.27 (0.41)** |
| A2 | 0.61 ( 0.65 ) | 0.56 ( 0.63 ) | 0.49 ( 0.57 ) | 0.71 ( 0.74 ) | 0.59 ( 0.65 ) | 0.17 ( 0.22 ) | 0.06 ( 0.08 ) | 0.02 ( 0.04 ) | 0.11 ( 0.15 ) |
| A3 | 0.61 (0.71) | 0.51 (0.70) | 0.49 (0.63) | 0.68 (0.78) | 0.57 (0.70) | 0.16 (0.32) | 0.06 (0.20) | 0.03 (0.16) | 0.11 (0.24) |
| A4 | 0.69 ( 0.75 ) | 0.65 ( 0.75 ) | 0.59 ( 0.69 ) | 0.74 ( 0.80 ) | 0.67 ( 0.75 ) | 0.30 ( 0.43 ) | 0.21 ( 0.35 ) | 0.19 ( 0.33 ) | 0.23 ( 0.36 ) |

Table 1: Performance of QA pipeline with different LLMs on a subset of test set. Since A2 and A4 depends on an OIE tool, we used red color to highlight the where IndIE tool performed better than Gen2OIE and green color to highlight where Gen2OIE performed better than IndIE. We used F1 score of ROUGE-1 (R1), ROUGE-2 (R2), ROUGE-3 (R3), and ROUGE-LCS (RL). Numbers written inside parenthesis represents the performance of $QA_{finetuned}$ whereas numbers written outside represents the performance of $QA_{base}$. Highest numbers in a column are highlighted in bold. It can be seen that A1 has consistently performed better than others.

| Pipeline | Approach | gemma-2b-it | gemma-7b-it | llama3.1-8b-it |
|---|---|---|---|---|
| $QA_{base}$ | B | 11GB x2<br>4 secs/question | 11GB x2<br>30 secs/question | 12GB x2<br>70 secs/question |
| | A1 | 3.5GB x2<br>4 secs/question | 6GB x2<br>8 secs/question | 7GB x2<br>58 secs/question |
| $QA_{finetuned}$ | B | 20GB x2<br>13 secs/question | 29GB x2<br>30 secs/question | 30GB x2<br>45 secs/question |
| | A1 | 8GB x2<br>13 secs/question | 21GB x2<br>20 secs/question | 23GB x2<br>22 secs/question |

Table 2: Computation footprint of QA pipeline with baseline (B) and A1 approaches. Memory consumption across two GPU cards is indicated by ‘x2’ symbol. It can be seen that A1 needs less compute and it is faster than B because of the short-context obtained from retriever.

```
Given the following question, you are given a ground-truth answer and two options.
Choose the option that is closest to the ground truth. You are only allowed to choose
one option. Print either “option1" or “option2". Print nothing else.
##Question
{question}
##Ground_Truth: {ground_truth}
##Option1: {option1}
##Option2: {option2}
```

Figure 4: Prompt used to query ChatGPT model to perform qualitative analysis.

more likely to contain intact coreferring mentions, and hence the resulting clusters of verbalized triples are shown to be giving better results.

After comparing the performance gain of A1 over B across three LLMs, we observed that the average percentage improvement of STS-MuTe and ROUGE scores was 4% and 47%, respectively for $QA_{base}$ whereas the improvement was 2% on both the metrics for $QA_{finetuned}$.

In order to perform a qualitative analysis on the answers generated by QA pipelines with B and A1, we decided to use LLM-as-a-judge. Since human annotators are time intensive and costly, LLM-as-a-judge has been used in many prior works (Pan et al., 2024; Chen et al., 2024). Our preliminary experiments on a subset of test set showed that base `gpt-4o-mini` from (Hurst et al., 2024) is able to outperform all $QA_{finetuned}$ on all the quantitative evaluation metrics. We used the prompt shown in Figure 3. Therefore, we decided to use ChatGPT LLM as a judge. We used `gpt-4o` to evaluate the answers generated from $QA_{finetuned}$ pipeline with all three LLMs. The prompt used for this task is shown in Figure 4. The percentage of questions where answers generated by A1 are preferred over the answers generated by B are as follows: 73% for gemma-2b-it, 93% for gemma-7b-it, and 51% for llama3.1-8b-it. Thus illustrating the better qualitative performance of A1 over B.

As an ablation of A1 approach, instead of fine-tuned APS model from (Mishra et al., 2024b) we also used LangChain[3] vectorstore retriever with

[3] https://github.com/langchain-ai/langchain

Huggingface embeddings (Wolf et al., 2020) from the `paraphrase-multilingual-MiniLM-L12-v2` model (Reimers and Gurevych, 2019). It was observed that A1 approach with fine-tuned APS model outperformed the LangChain variant of A1. We also used BM25 to select answer paragraphs for a given query from the corresponding context. It was observed that A1 approach with fine-tuned APS model ($A1_{APS}$) outperformed the LangChain variant and BM25 variant of A1 on all our evaluation metrics. Notably, 75% of the top-k paragraphs selected by $A1_{APS}$ differed from those retrieved by BM25, suggesting that, unlike BM25, the APS model captures information beyond bag-of-words features and probabilistic ranking.

The A1 approach does not only shows better performance than B on evaluation metrics but it is also less resource intensive. Table 2 shows the extent to which QA pipeline with A1 approach offers compute and time efficiency over the baseline (B). The QA pipeline with A4 using GenOIE tool has shown to be performing competitive to the best performing approach A1. However, it adds a compute overhead to the QA pipeline. The coref component takes 5GB on GPU and needs 0.14 seconds per article whereas Gen2OIE component takes 6.5GB on GPU and needs 15 seconds per article.

The resource intensive nature of B led to Out of Memory (OOM) errors on many test examples with the GPU cards we had at our disposal. Therefore, we compared B and A1 on the test examples common to both of them among the 1100 test examples. Table 3 highlights the performance of A1 over B on the common test examples. It shows A1 outperforming B in majority of the test examples most of the time. Among all test examples where A1 successfully generated an answer but B resulted in OOM error, the average number of tokens in the (long) context was 1238, 1189, and 1069 for QA pipelines with gemma-2b-it, gemma-7b-it, and llama3.1-8b-it, respectively.

To assess the performance of B relative to $A1_{APS}$, a subset of non-factoid QA pairs from the NaturalQA dataset (Kwiatkowski et al., 2019) was automatically translated into Hindi, Tamil, Telugu, Urdu, Marathi, and Bengali using the NLLB 1.3B model (Team et al., 2022). The selection focused on questions for which no short answer was available. The resulting dataset featured an average context length exceeding 8000 tokens. All associated resources will be made publicly accessible. Empirical results demonstrate that (with the exception of $QA_{base}$ with Llama 3.1) $A1_{APS}$ consistently outperforms B across evaluation metrics, within the ChatGPT-as-a-judge framework, and in terms of memory efficiency across all LLMs.

| Pipeline | Eval. Metric | gemma-2b-it | gemma-7b-it | llama3.1-8b-it |
|---|---|---|---|---|
| $QA_{base}$ | STS-MuTe | **60.7%** | 48.4% | **59.1%** |
| | ROUGE | **64.9%** | **58.7%** | **60.6%** |
| $QA_{finetuned}$ | STS-MuTe | **65.8%** | 45.4% | 48.3% |
| | ROUGE | **61.8%** | 39.9% | 44.1% |

Table 3: Percentage of test examples common to A1 and B, for which the average ROUGE scores (R1, R2, R3, RL) and STS-MuTe for A1 surpass those of B. Instances where A1 demonstrated superior performance (>50%) compared to B are indicated in **bold**.

## 6 Discussion

Building upon the superior performance of A1 compared to the baseline (B), we sought to analyze the functioning of the APS model, which focuses on context shortening. To achieve this, we employed post-hoc explainability methods to derive rationales for the predictions made by the APS model. Specifically, we utilized LIME (Ribeiro et al., 2016b) and SHAP (Lundberg and Lee, 2017b) techniques,

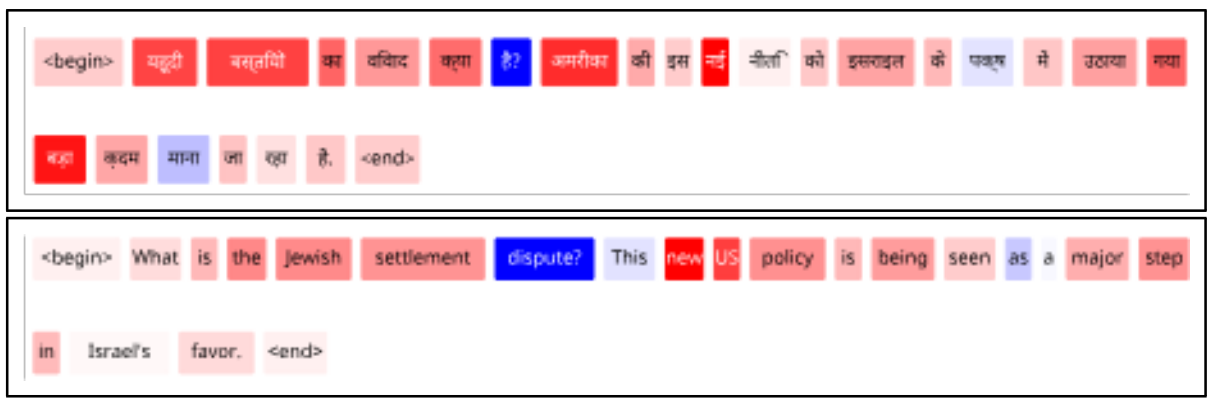

(a) LIME rationales where APS model score was 0.379 (left) and 0.377 (right).

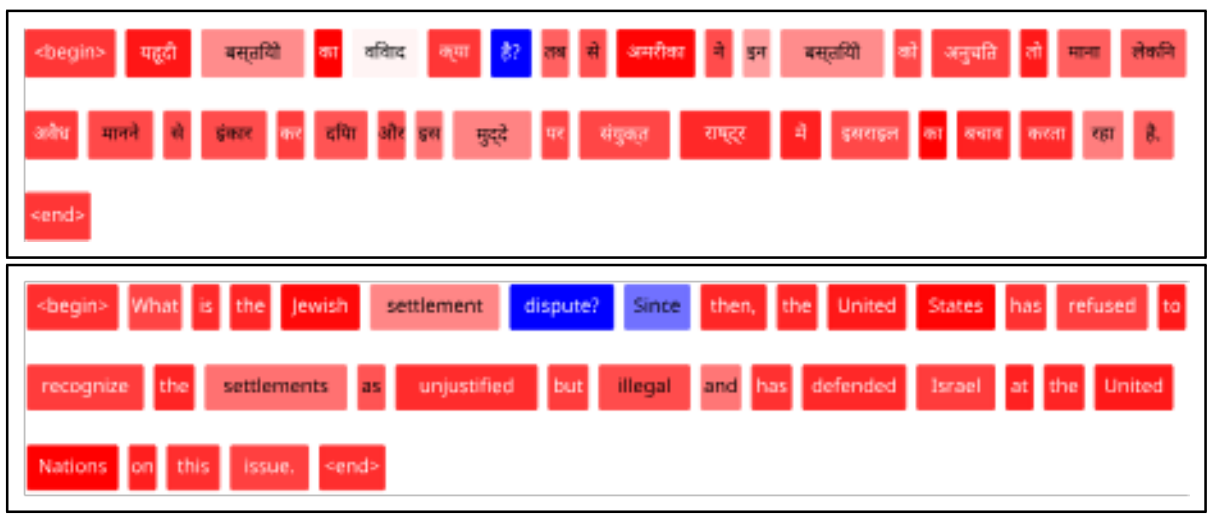

(b) LIME rationales where APS model score was 0.574 (left) and 0.56 (right).

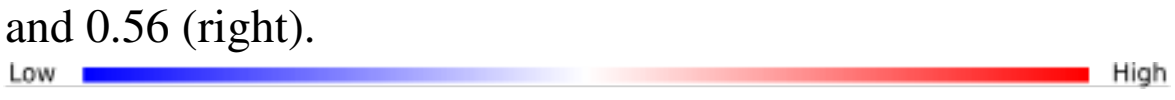

Figure 5: Rationales generated by LIME. The figures on the left correspond to the Hindi text from the test set, while those on the right display the English-translated text. Tokens with high relevance values (depicted in brighter red) indicate their significant contribution to the predicted logit. Masking these red-highlighted tokens is expected to reduce the model's confidence. Notably, as the APS model score increases, the rationales become more dispersed (shown in red). Similar patterns were observed with SHAP rationales.

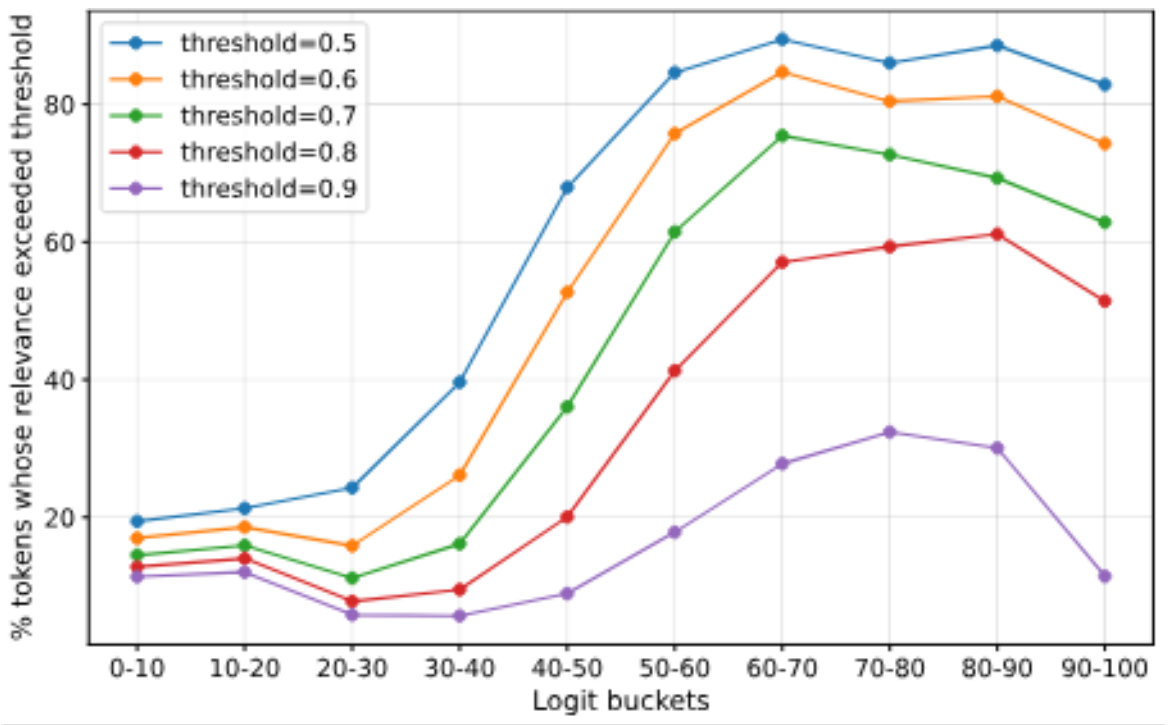

(a) LIME rationales.

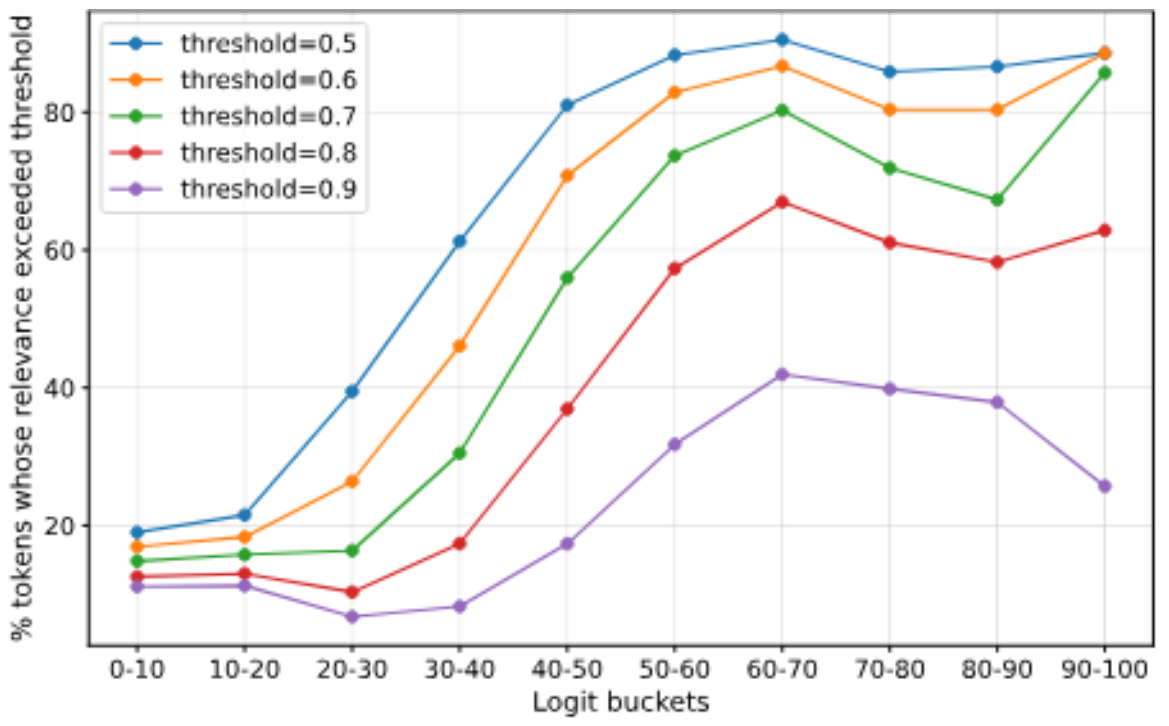

(b) SHAP rationales.

Figure 6: Consider an example to interpret the figure above: an APS model predicts a logit of 0.35 for a text containing 10 tokens, where 7 tokens exhibit a relevance value exceeding a specified threshold (e.g., 0.5). In this case, the logit bucket would fall within the range of 30–40, and the percentage of tokens with relevance values surpassing the threshold (0.5) would be 70%. The figure illustrates that higher logit values are positively correlated with a broader distribution of relevance values across varying thresholds.

facilitated through the Ferret library (Attanasio et al., 2023), to generate these rationales. Figure 5 presents a visualization of a sample rationale produced for a Hindi text, accompanied by an English translation automatically generated using the NLLB 1.3 billion model (Team et al., 2022).

An APS model assigns a high score when it confidently determines that the paragraph concatenated with the question is likely to address the given query. It was observed that a high relevance value (denoted by red color) was attributed to a significant number of tokens when the APS model predicted a high score, whereas fewer tokens were assigned high relevance values when the model predicted a low score. To substantiate this observation, an empirical validation was conducted.

The computation of rationales for an APS model prediction was observed to require between 4-15 seconds, depending on the length of the input text. Consequently, we performed rationale computation on a subset of the test set comprising over 8.5k question-paragraph pairs. The results, depicted in Figure 6, corroborate the aforementioned observations. These findings align with our intuition, indicating that only a minimal number of tokens are sufficient for the APS model to determine that a given paragraph does not answer the corresponding question. This observation reinforces the suitability of employing the APS model as a retriever within the QA pipeline.

| **QA pipeline** | | **Question Categories** | | | | |
|---|---|---|---|---|---|---|
| | | **Factoid** | **Evidence-Based** | **Debate** | **Reason** | **Experience** |
| gemma-2b-it | B | 30 vs 28 ( 32 ) | 26 vs 25 ( 32 ) | 16 vs 21 ( 13 ) | 12 vs 15 ( 10 ) | 3.6 vs 3 ( 2 ) |
| | A1 | 31 vs 32 ( 40 ) | 25 vs 28 ( 28 ) | 20 vs 15 ( 18 ) | 11 vs 11 ( 4 ) | 3.9 vs 3 ( 2 ) |
| gemma-7b-it | B | 30 vs 42 ( 40 ) | 25 vs 30 ( 29 ) | 19 vs 11 ( 17 ) | 11 vs 8 ( 9 ) | 4 vs 2 ( 2 ) |
| | A1 | 31 vs 25 ( 34 ) | 25 vs 31 ( 32 ) | 20 vs 22 ( 19 ) | 11 vs 4 ( 4 ) | 3.9 vs 6 ( 2 ) |
| llama3.1-8b-it | B | 31 vs 41 ( 40 ) | 25 vs 29 ( 22 ) | 20 vs 18 ( 24 ) | 11 vs 5 ( 6 ) | 3.9 vs 1 ( 3 ) |
| | A1 | 31 vs 46 ( 37 ) | 25 vs 24 ( 26 ) | 20 vs 11 ( 19 ) | 11 vs 5 ( 7 ) | 3.9 vs 4 ( 4 ) |

Table 4: QA pipelines across various question categories. The format employed in the cells is as follows: "*Total* vs *Best_100 for* $QA_{base}$ (*Best_100 for* $QA_{finetuned}$)" where *Total* and *Best_100* denote the percentage representation of a category among all generated answers and the best-100 answers, respectively. We selected best-100 answers by sorting the generated answers based on their average scores and selecting the top-100. If the *Best_100* percentage surpasses its corresponding *Total* percentage then it is highlighted in green ; otherwise, it is highlighted in red . It is evident that Factoid and Evidence-Based questions are more frequently observed in *Best_100* as compared to questions belonging to the Debate, Reason, and Experience categories.

The MuNfQuAD (Mishra et al., 2024b) includes question categories corresponding to each question, and its authors have demonstrated that while the majority of questions in MuNfQuAD are non-factoid, factoid questions are also represented within the dataset. We analyzed the patterns in question categories among the top 100 test examples. These examples were selected by sorting the generated answers based on their average scores and extracting the first 100 instances. The average scores were computed by taking an overall mean of semantic-level (STS-MuTe) and token-level (ROUGE) evaluation metrics. Consistent with the findings of (Bolotova et al., 2022), Table 4 demonstrates that the highest-quality answers are more frequently associated with factoid and evidence-based questions, as opposed to questions categorized under debate, reason, and experience.

## 7 Conclusion

Question Answering (QA), one of the earliest tasks in NLP, has seen the development of numerous approaches to address the task. With the emergence of Large Language Models (LLMs) and their superior performance across various NLP tasks, the focus of QA has shifted toward overcoming the limitations of LLMs, particularly the restricted context window, as well as addressing the relatively underexplored area of non-factoid question answering in low-resource languages. This study investigates the use of a retriever mechanism to shorten the context associated with a question. We explore several techniques, including Answer Paragraph Selection (APS), Open Information Extraction (OIE), and coreference resolution (coref), to implement a question-specific retriever. As a baseline, the entire context is provided to the LLM without any context reduction to answer the posed question. To evaluate the quality of the generated answers, we propose a Semantic Text Similarity score for Multilingual Text (STS-MuTe). Our findings indicate that the APS-based approach (A1) outperforms the baseline and other methods on both semantic and token-level metrics. Moreover, the A1 approach is found to be less resource-intensive. Additionally, experiments with post-hoc explainability methods show that the APS model aligns with human intuition when generating a score for a given question-paragraph pair.

## 8 Limitations

The Gen2OIE method was assessed on two Indic languages: Hindi and Telugu. In this study, we utilized the checkpoints for these languages and performed zero-shot queries for Urdu and Tamil, respectively. This approach was motivated by the substantial linguistic relatedness between Hindi-Urdu and Telugu-Tamil, as demonstrated by the online tool[4] from (Beaufils and Tomin, 2021).

The ROUGE Python package[5] employed in this study is designed specifically for the English language. To the best of our knowledge, no ROUGE package is currently available that supports the processing of multilingual texts written in non-Roman scripts. Consequently, we transcribed all the texts into Roman script using the method outlined in (Bhat et al., 2015). The results presented here may differ when using a ROUGE package specifically designed for languages with non-Roman scripts, particularly with respect to stopword removal.

Our QA pipeline utilizes pretrained LLMs. As noted by Ahuja et al. (2023), these LLMs are trained on extensive publicly available data, such as the Common Crawl dump. Consequently, there is a potential risk of inflated numbers in evaluation metrics. To investigate this, we excluded test data that was publicly available prior to the release of each model checkpoint[6]. This filtering resulted in 78, 197, and 18 examples for gemma-2b-it, gemma-7b-it, and llama3.1-8b-it, respectively. Compared to the results in Table 1, the reduction in evaluation metrics for $QA_{base}$ with the baseline (B) was 13%, 2.5%, and 22% for the QA pipelines employing these three LLMs, respectively. This finding indicates a slight decline in performance of QA pipeline when evaluated on data unseen during the pretraining of underlying LLM.

We acknowledge that the context size utilized in this study is considerably smaller than the token limits of many widely used LLMs. For example, the Gemma models support a token limit of 8K, whereas LLaMA 3.1 offers a token limit of 131K [7]. However, in the absence of a multilingual non-factoid QuAD dataset containing sufficiently long contexts, we assert that MuNfQuAD is the most suitable dataset for conducting experiments. One could argue that concatenating the context of a given question with the context of other questions might synthetically extend the context length. However, prior research has shown that the answer to a non-factoid question is typically localized within a small cluster in the document, rather than being dispersed across the entire text (Yang et al., 2016; Keikha et al., 2014). Therefore, we did not explore the potential benefits of artificially increasing the context length. While the retrieval-based pipeline employed in this study is theoretically capable of handling contexts of any length, its effectiveness in doing so should be evaluated on multilingual QuAD datasets with contexts exceeding the limits of LLMs.

[4] `http://www.elinguistics.net/Compare_Languages.aspx`

[5] `https://pypi.org/project/rouge-score/`

[6] We refereed to the last commit date of the model file in their huggingface repository.

[7] The token limit of an LLM is derived from the `max_position_embeddings` attribute in the model's `config.json` file available on its Huggingface repository.

## 9 Future Work

A promising direction for future work involves evaluating the proposed pipeline on multilingual datasets featuring longer contexts than those found in MuNfQuAD. Extracting text from school textbooks has been demonstrated as an effective approach for curating high-quality datasets in low-resource languages (Anand et al., 2024d,b,a,c). Furthermore, the APS model can serve as a reward model to align LLMs for generating more informative answers, facilitated by differential performance preference tuning algorithms (Li et al., 2024b; Naseem et al., 2024).

## Acknowledgments

Ritwik Mishra acknowledges with gratitude the partial financial support provided by the University Grants Commission (UGC) of India through the UGC Senior Research Fellowship (SRF) program. Rajiv Ratn Shah extends his appreciation for the partial funding received from the Infosys Center for AI, the Center of Design and New Media, and the Center of Excellence in Healthcare at IIIT Delhi.